%% file: acl_main.tex
\documentclass[11ptm]{article}

\usepackage[preprint]{latex/acl}

\usepackage{times}
\usepackage{latexsym}
\usepackage{tcolorbox}
\usepackage{listings}

\usepackage[T1]{fontenc}
\usepackage{multirow}
\usepackage{multicol}
\usepackage{graphicx}

\usepackage[utf8]{inputenc}

\usepackage{microtype}

\usepackage{inconsolata}

\usepackage{graphicx}
\usepackage{amsmath}
\usepackage{booktabs}
\usepackage{pifont}
\usepackage{xcolor}
\usepackage[colorinlistoftodos,textsize=small]{todonotes}

\newcommand{\cmark}{\ding{51}}
\newcommand{\xmark}{\ding{55}}

\title{Self-Evolving Visual Questioner}

\author{
Yijun Liang$^{1}$ \quad
Hengguang Zhou$^{2}$ \quad
Ming Li$^{1}$ \quad
Lichen Li$^{3}$ \quad
Cho-Jui Hsieh$^{2,4}$ \quad
Tianyi Zhou$^{5}$ \\[0.5em]
$^{1}$University of Maryland, College Park \quad
$^{2}$University of California, Los Angeles \\
$^{3}$Peking University \quad
$^{4}$Arena \quad
$^{5}$MBZUAI \\[0.5em]
}

\begin{document}
\maketitle

\input{Section/abstract}

\input{Section/introduction}

\input{Section/method}
\input{Section/experiment}

\input{Section/related_work}
\input{Section/conclusion}

\bibliography{custom}

\clearpage
\newpage
\appendix
\input{Section/Appendix/detailed_qg_evaluation}
\input{Section/Appendix/Prompts}
\input{Section/Appendix/Training_parameter}
\input{Section/Appendix/detailed_qa_performance}
\input{Section/Appendix/human_alignment}
\input{Section/Appendix/ExistingVQG_Performance}
\input{Section/Appendix/Cases}

\end{document}

%% file: Section/abstract.tex
\begin{abstract}
Vision-language models (VLMs) are typically trained as passive answerers, while their ability to actively ask diverse, non-trivial, visual-centric and grounded questions remains underexplored. Existing visual questioners' performance is bottlenecked by the availability of high-quality training data or the cost of curating them. 
We show that a VLM can continuously improve itself as a visual questioner 
without any external supervision. 
We propose a self-evolving framework that uses a VLM itself as both a proposer and a filter to produce harder, more informative, and visual-centric questions, while maintaining their exploration diversity to avoid training collapse. These questions are then used to train the VLM in both questioner and answerer modes. 
To evaluate the questioner, we introduce an agentic protocol that assesses questions along perception, reasoning, and diversity dimensions. 
Experiments across various backbone VLMs show that our method substantially enhances the quality and substantially expands the difficulty boundary of autonomous question generation. 
Under the same budget, our self-supervision is more effective than training on the static source data. Moreover, the self-evolving questioner remains a competitive or even better answerer
\footnote{Project is available at this \href{https://joliang17.github.io/SelfEvolvingVQG/}{link}.}. 
\looseness-1
\end{abstract}

%% file: Section/introduction.tex
\section{Introduction}
\label{sec:introduction}


Vision-language models (VLMs) have achieved substantial progress on visual question answering, multimodal instruction following, and visual reasoning benchmarks. 
However, most existing VLMs are trained and evaluated primarily as passive answerers: questions are treated as fixed inputs provided by humans, curated datasets, or stronger external models, and learning focuses on producing better answers~\citep{liu2023visual,wu2024v,zhang2026flatland}. 
This leaves the complementary ability to ask informative and visually grounded questions comparatively underexplored. 
We view visual questioning as a fundamental capability for intelligent systems: it reflects whether a model can actively inspect an image, identify informative visual evidence, and formulate questions that require meaningful perception and reasoning. 
Since this capability is expressed through the questions a model generates, question quality provides a concrete lens for studying how well a model can inspect image content, ground visual evidence, and pose reasoning-oriented queries. 
Better visual questions can further expose richer evidence and reasoning paths, making them valuable supervision for improving visual questioning itself.


Despite its importance, existing visual question generation (VQG) methods remain fundamentally bottlenecked by static data distributions~\citep{mi2024convqg,shen2024ask}.
Many approaches rely on human annotations, curated datasets, or stronger external models to provide supervision and consequently learn primarily to imitate existing question patterns~\citep{zhao2024lova3,xie2025explicitly,zhang2025diagram}. 
As a result, the diversity, grounding quality, and reasoning complexity of generated questions are largely bounded by the underlying data distribution. 
Generated questions often concentrate on repetitive templates, salient objects, and surface-level recognition, and may not require careful inspection of image-specific visual evidence. 
This leads to shallow image exploration and limited perceptual reasoning.
To sustain improvement, supervision that can keep expanding toward broader, harder, and more visually grounded questions are needed. 
However, continuously collecting such question annotations from humans or stronger external models is expensive, difficult to scale, and challenging to control.

To move beyond this bottleneck, a natural alternative is to allow the model itself to autonomously construct better supervision through self-generated questions. 
However, simple scaling self-generated supervision does not automatically lead to self-improvement. 
Without mechanisms that explicitly preserve exploration diversity, improve difficulty, and ensure visual grounding, the model may reinforce its own biases and collapse toward repetitive, low-information, or weakly grounded questions. 
Consequently, the key challenge is not merely generating more questions, but sustaining the continual evolution of questioning capability while progressively expanding diversity and reasoning difficulty throughout training.

To address this challenge, we propose a self-evolving visual questioner framework that enables a VLM to improve itself as a visual questioner, without human annotations, external teacher models, or auxiliary reward models. 
Rather than relying on continuously collected datasets, our framework uses the current model itself to iteratively construct progressively stronger supervision from unlabeled images. 
Given an image, the model first proposes candidate questions under multiple visual intents to encourage broader coverage of the space for all potential questions. 
It then applies a rewrite and filtering procedure to select question-answer candidates that are more visually grounded, more informative, and more reasoning-oriented while preserving diversity across question types. 
Unlike conventional self-training, which largely reinforces existing prediction distributions and risks collapse toward narrow behaviors, our framework is inherently exploratory, continually evolving the question distribution toward greater diversity, stronger visual grounding, and richer reasoning complexity. 
The selected data are used for dual-format training with both QG-format and QA-format supervision, enabling the model to improve its questioning behavior while maintaining its answering ability. 
The trained model is then reused as the proposer in the next round, forming a closed-loop self-improvement framework for visual questioning that can produce increasingly diverse, difficult, and visual-centric supervision from the same unlabeled image pool over iterative rounds.

To evaluate visual questioning capability beyond conventional QA accuracy, we introduce an agentic 
evaluation protocol that assesses generated questions along perception, reasoning, and diversity dimensions. 
Experiments across multiple VLM backbones show that our self-evolving framework substantially improves the visual-centric quality of generated questions while largely preserving QA performance. 
After two self-evolving rounds, \textbf{the average QG score increases by $\sim$82\%} relative to the initial base model.
The generated questions become better at searching for informative visual evidence and covering broader image content, while also showing stronger contextual and spatial reasoning. 
In particular, our proposal-and-filter mechanism effectively maintains exploration diversity and visual-centric reasoning throughout self-evolution, mitigating the degeneration toward repetitive and shallow questioning behaviors commonly observed in static or unconstrained self-generated supervision.

While our primary goal is not to optimize downstream VQA performance, QA accuracy remains stable and even improves on several benchmarks, suggesting that the model becomes a stronger visual questioner without sacrificing its answering capability. 
Ablations on supervision format, supervision source, and filtering strategy further show that our generated data provides a more effective small-budget training signal than directly sampled source annotations. 
These results suggest that controlled self-evolution can produce increasingly diverse, difficult, and visually grounded questions, providing more informative supervision for VLM training without relying on continuously collected external annotations.\looseness-1

\textbf{Our contributions} are summarized as follows:
\begin{itemize}
    \item We introduce a fully autonomous self-evolving framework for visual question generation, where a VLM continuously improves its questioning capability using only self-generated training data. 
    \item We propose a multi-round generating, rewriting, and filtering mechanism that promotes diversity, difficulty, and visual grounding while preventing degeneration during recursive self-training.
    \item We introduce a structured agentic evaluation protocol that measures visual-centric questioning quality across perception, reasoning, and diversity dimensions, enabling fine-grained analysis beyond aggregate QA accuracy.
    \item We conduct extensive experiments and ablations across multiple models, text, and image sources, and filtering strategies, demonstrating the effectiveness and robustness of our framework.
\end{itemize}

%% file: Section/method.tex
\section{From Answerer to Questioner}
\label{sec:method}

\input{Figure_tex/pipeline}

\subsection{Overview}
\label{sec:method_overview}

We study whether a vision-language model (VLM) can improve its visual question generation (QG) ability from its own generated data while preserving downstream visual question answering (QA) performance. 
Given an image collection $\mathcal{I}$ and an initial model $M_0$, our goal is to construct visual-centric question-answer data that improves the model's ability to generate diverse, grounded, and reasoning-oriented questions.

Our framework consists of two stages. 
First, the model proposes candidate questions, rewrites them into harder visual-centric questions, and filters candidates based on answerability, visual grounding, and perception/reasoning difficulty. 
Second, the retained question-answer pairs are used for dual-format training: QA-format supervision anchors answering behavior, while QG-format supervision teaches the model to generate both questions and answers. 
The adapted model can be reused as the new question proposer int he next round, forming an iterative self-improvement loop over the same unlabeled image pool.
For analysis, we introduce an agentic visual questioning capability evaluation protocol that measures generated questions along perception, reasoning, and diversity dimensions.

\subsection{Problem Formulation}
\label{sec:method_formulation}

A visual QA example consists of an image $x$, a question $q$, and an answer $a$. 
Standard VQA training optimizes the conditional answer distribution:

\begin{equation}
    p_\theta(a \mid x,q).
\end{equation}

In contrast, visual question generation requires the model to produce an informative question conditioned on the image and control information. We formulate QG as learning:\looseness-1

\begin{equation}
    p_\theta(q \mid x,c),
\end{equation}
where $c$ denotes optional conditioning information such as a target question type, answer style, or reasoning intent. 
In our framework, such intents encourage broader coverage of the per-image question space; when no control is provided, the model generates questions directly from the image.




\subsection{Agentic Questioning Capability Evaluation Protocol}
\label{sec:method_qg_eval}

A central challenge in evaluating visual questions is that surface-level quality does not necessarily reflect visual informativeness. 
A question may be well-formed but shallow, ambiguous, weakly grounded in the image, or redundant with other questions for the same image. 
We therefore design an evaluation protocol to assess whether generated questions are visual-centric, challenging, and diverse.

Our protocol evaluates generated questions at two levels. 
At the individual-question level, we measure \textbf{perception difficulty} and \textbf{reasoning difficulty}. 
\textbf{Perception difficulty} concerns what visual evidence must be identified to answer the question, and consists of two dimensions: \textit{Visual Search Difficulty}, which measures how difficult it is to locate the required evidence, and \textit{Visual Evidence Coverage}, which measures how broadly the question depends on evidence across the image. 
\textbf{Reasoning difficulty} concerns how the identified evidence must be interpreted, and consists of two dimensions: \textit{Visual Context Reasoning}, which measures contextual interpretation from visible cues, and \textit{Visual Spatial Reasoning}, which measures reasoning over spatial relations among image elements. 

At the question-set level, we measure \textbf{diversity} through \textit{Questioning Diversity}, which captures redundancy among questions generated for the same image. 
This set-level dimension is important because a question can be valid and reasoning-oriented in isolation, yet provide little additional supervision if it is a near-duplicate of other questions for the same image.


For individual-question evaluation, we use a GPT-based judge with clearly defined rubrics for the four perception and reasoning dimensions. 
The judge scores each question conditioned on the image, allowing the evaluation to focus on visual evidence and reasoning requirements rather than surface-level wording.
For question-set diversity, we use a sentence embedding model to compute semantic distances among questions generated for the same image, measuring whether the question set provides non-redundant visual supervision.
Detailed rubrics and diversity computation are provided in Appendix~\ref{app:detail_QG_definition}.


\section{Self-Evolving Visual Questioner}
\label{sec:method_data_generation}
Our framework consists of two stages: self-supervision data construction and dual-format model training, aiming to improve the VLM's questioning capability while preserving its answering ability. 
Given unlabeled images, the same VLM is used throughout the data-generation process under different roles: it proposes candidate questions, rewrites them into harder and more visual-centric questions, and filters the candidates based on visual answerability and perception/reasoning difficulty. 
This refinement prevents the raw proposals from being used directly for training and mitigates collapse toward repetitive or low-information questions.
\textbf{Importantly, this rewriting-and-filtering procedure is used only during self-evolution training. }
During evaluation, we directly assess the model’s unmodified generated questions without additional rewriting or filtering, ensuring that the reported results reflect the model’s intrinsic questioning capability.
The resulting self-generated and self-refined supervision requires neither external teachers nor human annotations, since it is generated, improved, and selected by the model itself. 
The retained question-answer pairs are then used for dual-format training, enabling the model to improve question generation while maintaining its answering ability.\looseness-1

\subsection{Question Proposal}
\label{sec:method_seed_generation}

For each image $x_i\in\mathcal{I}$, the current model $M_t$ first proposes a set of candidate questions:
\begin{equation}
    \mathcal{Q}^{\mathrm{prop}}_i = \{q^{\mathrm{prop}}_{i,1}, \ldots, q^{\mathrm{prop}}_{i,K}\}.
\end{equation}
Here $M_t$ can be either the initial model $M_0$ or a model adapted in a previous self-evolution round. 
The proposal prompt asks the model to generate questions that are answerable from the image and grounded in visible evidence. The answers are then generated by $M_t$ based on image $x_i$ and proposal $q^{\mathrm{prop}}_{i}$.

To construct a diverse proposal pool, we prompt the model with multiple question intents, such as direct recognition, comparison, spatial relations, scene understanding, and grounded reasoning. 
These intents encourage the model to cover different aspects of the image and reduce the tendency to produce shallow or repetitive question templates.

For each proposed question, we use the same model to generate the corresponding answer, forming an initial set of question-answer proposal triples:
\begin{equation}
    \mathcal{C}^{\mathrm{prop}}_t = \{(x_i, q^{\mathrm{prop}}_{i,j}, a^{\mathrm{prop}}_{i,j})\}_{i,j}.
\end{equation}
These proposal triples serve as the starting candidates for the subsequent rewrite and filtering stages.

\subsection{Question Rewriting}
\label{sec:method_self_evolution}
A key design choice in our framework is to separate question proposal from question rewriting. 
The current checkpoint $M_t$ serves as the evolving proposer: as it is trained across rounds, it becomes increasingly specialized in direct visual question generation and changes proposal distribution over the image collection. 
This changing distribution allows later rounds to explore new candidate supervision.
The rewriting step then refineds these proposals along specified visual-centric directions, such as increasing visual inspection difficulty, evidence grounding, contextual reasoning, or spatial reasoning.

However, using the same adapted checkpoint to rewrite its own proposals would tightly couple proposal and refinement, causing both steps to follow the same generation distribution.
We therefore instantiate rewriting with the initial checkpoint $M_0$, which acts as the model's pre-adaptation question-evolution operator. 
This introduces a different perspective into the data-construction process, adding useful exploration and diversity without external supervision.
In this design, $M_t$ determines what questions are explored, while $M_0$ adds variation to the rewriting process.

Given an image $x$, a question proposal $q^{\mathrm{prop}}$, and an evolution instruction $r$, the rewriting operator produces an evolved question:
\begin{equation}
    q^{\mathrm{rw}} \sim M_0(\cdot \mid x, q^{\mathrm{prop}}, r).
\end{equation}
Instruction $r$ specifies the intended difficulty direction, e.g., increasing visual inspection difficulty, evidence grounding, contextual reasoning, or spatial reasoning.


The rewritten question is paired with answer $a^{\mathrm{rw}}$ generated by $M_0$ conditioned on image $x$ and rewritten question $q^{\mathrm{rw}}$:
This forms a rewritten candidate:
\begin{equation}
    \mathcal{C}^{\mathrm{rw}}_t = \{(x_i, q^{\mathrm{rw}}_{i,j}, a^{\mathrm{rw}}_{i,j})\}_{i,j}.
\end{equation}
The resulting rewritten candidates are then passed to the filtering stage in Section~\ref{sec:method_self_filtering}.

\subsection{Question Filtering}
\label{sec:method_self_filtering}

The rewritten candidate pool $\mathcal{C}^{\mathrm{rw}}_t$ contains questions with varying validity, specificity, and difficulty. 
Before training, we therefore apply a model-based filtering step to retain only useful supervision.
For each rewritten candidate  $(x,q^{\mathrm{rw}},a^{\mathrm{rw}})$, the model compares it with its corresponding proposal triple $(x,q^{\mathrm{prop}},a^{\mathrm{prop}})$ and checks whether the rewritten QA pair remains visually answerable, grounded in image, and more challenging than the original proposal.

The filtering rule retains candidates that are visually valid and improve over their proposals in either perception or reasoning difficulty. 
Specifically, we discard candidates that are ambiguous, weakly grounded, visually unanswerable, or not improved over the original proposal. 
The retained candidates form the training dataset $\mathcal{D}^{(t)}$ used in the following training stage.

\subsection{Training and Iterative Self-Improvement}
\label{sec:training_iterative_self_improvement}

\subsubsection{Dual-Format Supervision}
\label{sec:method_dual_format}
After filtering process, the retained question-answer pairs are used to adapt the VLM. 
Our training stage mixes two supervision formats: QG-format supervision and QA-format supervision.
The QG-format presents only the image as input and trains the model to generate both the question and its corresponding answer, encouraging the model to generate visual-centric questions grounded in answerable image evidence.
The QA-format presents the image and question as input and trains the model to generate the answer, anchoring the model's answering behavior and helping preserve downstream VQA ability. 

Formally, given the retained dataset
\begin{equation}
    \mathcal{D}^{(t)} = \{(x_i, q_i, a_i)\}_{i=1}^{N},
\end{equation}
we construct two training set:
\begin{align}
    \mathcal{D}^{(t)}_{\mathrm{QG}}
    &= \{x_i \rightarrow (q_i, a_i)\}_{i=1}^{N}, \\
    \mathcal{D}^{(t)}_{\mathrm{QA}}
    &= \{(x_i, q_i) \rightarrow a_i\}_{i=1}^{N}.
\end{align}
The final training set is the mixture of these sets:
\begin{equation}
    \mathcal{D}^{(t)}_{\mathrm{train}}
    =
    \mathcal{D}^{(t)}_{\mathrm{QG}}
    \cup
    \mathcal{D}^{(t)}_{\mathrm{QA}}.
\end{equation}
We train the model with standard supervised fine-tuning on the target tokens in both formats. 
This improves question generation while maintaining the model's answering ability.

\subsubsection{Iterative Self-Improvement}
\label{sec:method_training}
Given $\mathcal{D}^{(t)}_{\mathrm{train}}$, we fine-tune the current model $M_t$ to obtain $M_{t+1}$. 
In the one-round setting, the initial model $M_0$ generates and filters candidate data, and the retained dataset is used to train $M_1$. 
For iterative self-improvement, the adapted model is reused as the proposer in the next round: producing a new set of question proposals. 
These proposals are then rewritten and filtered by $M_0$, and the resulting mixed-format supervision is used to train the next model.
This forms a self-improvement loop:\looseness-1
\begin{equation}
    M_t
    \rightarrow
    \mathcal{D}^{(t)}_{\mathrm{train}}
    \rightarrow
    M_{t+1}.
\end{equation}

Overall, our method improves QG through the interaction of rewriting, filtering, and dual-format training. 
Rewriting expands the candidate pool toward harder and more visual-centric questions, Filtering removes invalid or non-improving generations, and dual-format supervision improves question generation while anchoring answering behavior. 
Together, these components enable the model to improve visual question generation while maintaining competitive downstream QA performance.

%% file: Figure_tex/pipeline.tex
\begin{figure*}[t]
  \includegraphics[width=\linewidth]{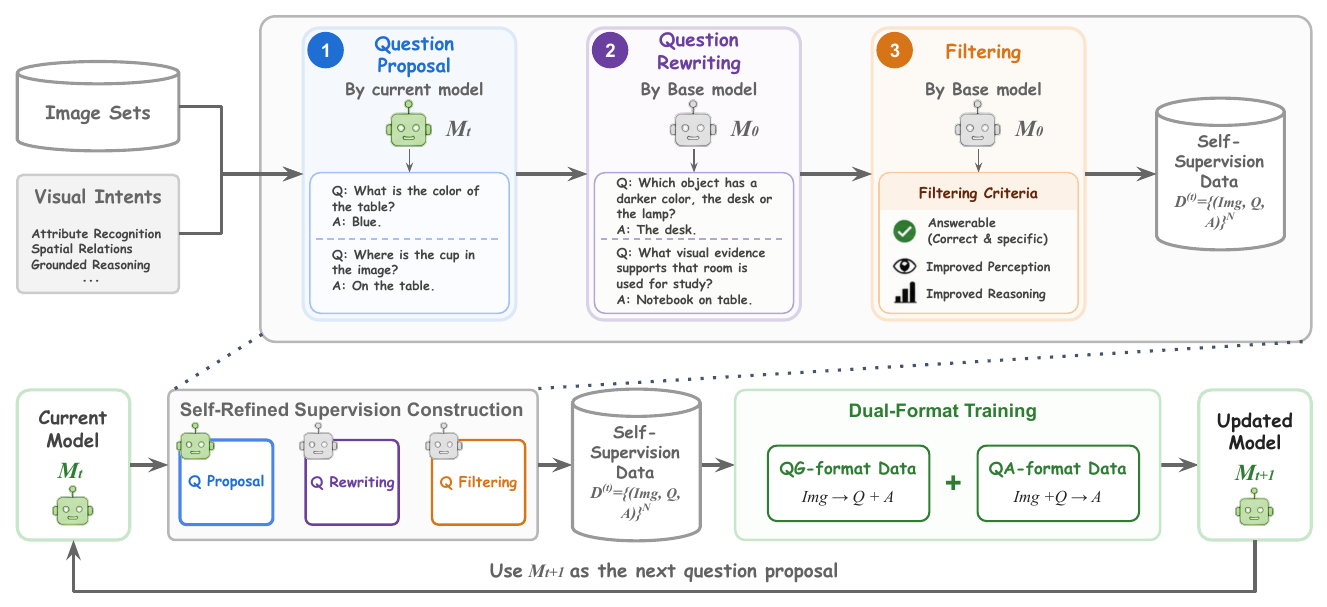}
    \caption{
    \textbf{Overview of our self-evolving visual questioner framework.}
    Given unlabeled image sets and visual-intent prompts, the current model $M_t$ first proposes candidate questions. 
    A stable base model $M_0$ then rewrites and filters these proposals to construct self-supervision data $\mathcal{D}^{(t)}$ with improved answerability, perception difficulty, and reasoning difficulty. 
    The refined supervision is used for dual-format training with both QG-format and QA-format objectives, producing an updated model $M_{t+1}$. 
    The updated model is reused as the proposer in the next round, forming an iterative self-improvement loop.
    }
  \label{fig:pipeline}
\end{figure*}

%% file: Section/experiment.tex
\section{Experiments}
\label{sec:exp}

In this section, we evaluate whether our self-evolving visual questioner framework improves question generation (QG) quality while preserving question answering (QA) performance.

Our experiments address four questions:
\begin{itemize}
    \item Can our self-evolving framework improve QG quality within a single round?
    \item Does iterative self-improvement further improve QG quality across rounds? 
    \item How do the effects vary across different generations and training settings?
    \item Does improved question quality provide more effective supervision for downstream QA training?
\end{itemize}

\subsection{Experimental Setup}
\label{sec:exp_setup}

We evaluate our self-evolving questioner framework on three VLM backbones: Qwen2.5-VL-3B-Instruct, Qwen2.5-VL-7B-Instruct~\citep{bai2025qwen25vltechnicalreport}, and Qwen3VL-4B-Instruct~\citep{bai2025qwen3}. 
Implementation details and training hyperparameters are provided in Appendix~\ref{app:training_parameters}.

We report results at three stages. 
\textbf{Base} denotes the original pretrained VLM. 
\textbf{First-round} denotes the model adapted with supervision constructed in the first round. 
\textbf{Second-round} denotes the model adapted with supervision constructed by reusing the first-round model as the question proposer. 
Unless otherwise specified, training uses $10$K examples and the QA+QG format described in Section~\ref{sec:training_iterative_self_improvement}.

To evaluate QG ability, each model generates $5$ questions per image using the same decoding parameters. We then apply the visual questioning capability evaluation protocol described in Section~\ref{sec:method_qg_eval}. 
For the four individual-question dimensions, we use GPT-5.4~\citep{singh2026openaigpt5card} as an image-conditioned judge. 
The judge receives the image and one generated question at a time, without access to the model identity. 
Each dimension is scored on a $0$--$5$ rubric and normalized to $[0,1]$. 
We first average scores over questions for each image and then average over images. 
Questioning diversity is computed within the generated question set for each image using Qwen3-Embedding-4B~\citep{zhang2025qwen3}.

Specifically, we evaluate QG on a held-out set of 100 diverse natural images sampled from CVBench. 
We report five dimensions: Visual Search Difficulty (\textbf{Search}), Visual Evidence Coverage (\textbf{Coverage}), Visual Context Reasoning (\textbf{Context}), Visual Spatial Reasoning (\textbf{Spatial}), and Questioning Diversity (\textbf{Diversity}). 
To evaluate downstream answering ability, we test models on CVBench~\citep{tong2024cambrian}, SparBench~\citep{zhang2026flatland}, VStarBench~\citep{wu2024v}, and RealWorldQA~\citep{xai2024realworldqa}. 
We report accuracy on each benchmark and the average accuracy across all benchmarks.\footnote{Detailed benchmark accuracies are provided in Appendix~\ref{tab:appendix_qa_details}.}

\subsection{Main Results}
\label{sec:exp_main_3b}

\input{Table/main_results_v2}

Table~\ref{tab:main_results} reports dimension-level QG evaluation results together with downstream QA performance.

\paragraph{Our framework improves questioning while preserving answering ability.}
According to Table~\ref{tab:main_results}, our self-evolving questioner framework consistently improves QG quality among all evaluated dimensions and model backbones. 
The gains are observed not only for Qwen2.5-VL models, but also for the stronger Qwen3VL-4B baseline, which already starts with relatively high QG scores. 
This suggests that the proposed generation, rewriting, and filtering process can further improve the model's ability to generate visual-centric, grounded, and reasoning-oriented questions even when the base model is already strong.
Importantly, these QG improvements do not come at the cost of downstream answering ability. 
Fine-tuning on a small-scale dataset can shift a VLM away from its original answering behavior, especially when the supervision focuses on a new generation format. 
Under our dual-format training setup, however, $10$K self-supervision examples are sufficient to improve QG quality without degrading downstream QA performance. 


\paragraph{Iterative self-improvement enables continued QG improvement.}
Beyond one-round adaptation, Table~\ref{tab:main_results} shows that QG quality continues to improve across rounds. 
The first round improves all five QG dimensions over the base model, and the second round further improves upon the first round. 
This suggests that the framework is not merely exploiting a one-time benefit from generated supervision; instead, the adapted model produces stronger question proposals in later rounds, which are further refined through rewriting and filtering.
By reusing the same unlabeled image pool, the framework progressively reshapes the generated question distribution toward broader coverage, stronger grounding, and higher perception/reasoning difficulty. 
These results support our central goal of improving visual question generation across rounds while maintaining competitive answering performance.

\subsection{Ablation Studies}
\label{sec:exp_ablation}

We next ablate the supervision format, supervision type, and filtering process to identify which components drive these improvements.

\subsubsection{Effect of Supervision Format}
\label{sec:ablation_qaiqa}

\input{Table/ablation_dataformat_v3}

Table~\ref{tab:ablation_supervision} studies the effect of supervision format while keeping the retained self-generated examples fixed.  
We compare QA-only training, QG-only training, and the proposed QA+QG dual-format training. 
QA-only trains the model to answer generated questions, QG-only trains the model to generate question-answer pairs from the image, and QA+QG combines both formats.

QG-only generally improves QG scores over QA-only, showing that explicit image-to-question-answer supervision strengthens question generation. 
However, QG-only can reduce downstream QA performance, indicating that training only in the QG format may shift the model away from its answering behavior. 
QA+QG provides the best overall trade-off: it improves QG average over QA-only across all three backbones, while achieving the best or comparable QA performance. 
These results support our dual-format design, where QG-format supervision improves visual question generation and QA-format supervision helps preserve downstream answering ability.

\subsubsection{Effect of Self-Evolution and Filtering}
\label{sec:exp_filtering_ablation}

\input{Table/ablation_generationprocess_v2}

Table~\ref{tab:ablation_filtering} ablates the effects of rewriting and filtering in Qwen2.5-VL-3B setting. 
Starting from direct self-generated supervision, the model already improves over the base model on visual search, evidence coverage, contextual reasoning, and spatial reasoning, showing that generated question-answer pairs can provide a useful signal for improving QG. 
Adding rewriting further strengthens all five QG dimensions, suggesting that rewriting expands the candidate pool toward harder and more visual-centric questions.

The full pipeline achieves the strongest QG performance across all dimensions. 
Compared with using rewriting alone, adding filtering improves visual search from 0.353 to 0.433, evidence coverage from 0.520 to 0.547, contextual reasoning from 0.507 to 0.553, spatial reasoning from 0.293 to 0.367, and diversity from 0.320 to 0.350. 
This indicates that filtering does more than remove invalid samples: it selects candidates that better match our desired QG properties, including stronger grounding, higher perception/reasoning difficulty, and lower redundancy. 
Together, these results show that rewriting and filtering play complementary roles: rewriting expands the candidate pool, while filtering selects higher-quality supervision for training.

\subsubsection{Effect of Self-Supervision Data}
\label{sec:ablation_data_source}

\input{Table/ablation_datasource_v2}

Table~\ref{tab:ablation_source} compares different sources of question-answer supervision under the same SAT image source, QA+QG training format, and training budget of $10$K examples. 
We compare directly sampled Original SAT annotations with our self-supervision data, which is constructed through the proposed rewriting and filtering process. 
Although the original SAT is a large-scale instruction-tuning dataset constructed from image-based rubrics, directly sampling a small subset of its annotations is not the most effective choice in this setting. 
In our setting, Original SAT improves visual search difficulty, but leads to lower average QA accuracy and weaker reasoning and diversity scores. 
This suggests that, under a limited training budget, source annotations may provide a less targeted signal for improving the desired QG properties.

In contrast, our self-supervision data consistently improves the overall QG profile while also slightly improving average QA performance. 
Compared with the original SAT, it achieves stronger scores on the majority of dimensions, indicating that the generated supervision is better aligned with producing more visual-centric and reasoning-demanding questions. 
The mixture of Original SAT and self-supervision data also improves over Original SAT alone, but remains weaker than using self-supervision data alone.
This suggests that naive mixing can dilute the targeted signal produced by our rewriting and filtering process. 
Overall, these results show that, under a fixed small-data budget, self-supervision data provides a more effective adaptation signal for improving QG quality while preserving QA performance.

\subsection{Do Better Questions Help Supervision?}
\label{sec:ablation_betterquestion}

\input{Table/ablation_betterquestion}
To examine whether improved question generation leads to more useful downstream supervision, we compare two training sets that differ only in the source of their generated questions. 
One uses questions generated by the base Qwen2.5-VL-3B model, while the other uses questions generated by the improved QG model. 
In both settings, we use GPT-5.4 to generate the corresponding answers, which controls for answer quality and isolates the effect of question quality.

As shown in Table~\ref{tab:ablation_gpt_answer}, training with questions from the improved QG model achieves higher average QA accuracy, improving from 61.90\% to 63.32\%. 
The gain is especially clear on CVBench-3D, where accuracy increases from 69.25\% to 75.58\%, and the improved-question setting also performs better on CVBench-2D, VStar, and RWQA. 
These results indicate that improving QG quality is valuable beyond the question-generation task itself: questions that require richer visual evidence and reasoning provide more informative supervision for downstream QA training, leading to more effective model adaptation.

%% file: Table/main_results_v2.tex
\begin{table*}[t]
    \centering
    \small
    \setlength{\tabcolsep}{3.2pt}
    \resizebox{0.9\textwidth}{!}{
    \begin{tabular}{l l ccccc c c}
    \toprule
    \multirow{2}{*}{\textbf{Backbone}} &
    \multirow{2}{*}{\textbf{Training stage}} &
    \multicolumn{5}{c}{\textbf{QG Dimension Scores}} &
    \multirow{2}{*}{\textbf{QG Avg.}} &
    \multirow{2}{*}{\textbf{QA Avg.}} \\
    \cmidrule(lr){3-7}
    & &
    \textbf{Search} &
    \textbf{Coverage} &
    \textbf{Context} &
    \textbf{Spatial} &
    \textbf{Diversity} &
    &
    \\
    \midrule
    Qwen2.5-VL-3B & Base
    & 0.26 & 0.39 & 0.30 & 0.03 & 0.28 & 0.25 & 60.49 \\
    Qwen2.5-VL-3B & First-round
    & 0.43 & 0.55 & 0.55 & 0.37 & 0.35 & 0.45 & 62.26 \\
    Qwen2.5-VL-3B & Second-round
    & \textbf{0.49} & \textbf{0.61} & \textbf{0.64} & \textbf{0.40} & \textbf{0.37} & \textbf{0.50} & \textbf{62.56} \\
    \midrule
    Qwen2.5-VL-7B & Base
    & 0.25 & 0.35 & 0.27 & 0.11 & 0.32 & 0.26 & 65.90 \\
    Qwen2.5-VL-7B & First-round
    & 0.46 & 0.57 & 0.59 & 0.35 & 0.38 & 0.47 & \textbf{67.02} \\
    Qwen2.5-VL-7B & Second-round
    & \textbf{0.51} & \textbf{0.59} & \textbf{0.64} & \textbf{0.40} & \textbf{0.40} & \textbf{0.51} & 66.90 \\
    \midrule
    Qwen3VL-4B & Base
    & 0.38 & 0.51 & 0.45 & 0.18 & 0.26 & 0.36 & \textbf{71.31} \\
    Qwen3VL-4B & First-round
    & 0.52 & 0.63 & 0.62 & 0.54 & 0.38 & 0.54 & 71.12 \\
    Qwen3VL-4B & Second-round
    & \textbf{0.56} & \textbf{0.64} & \textbf{0.65} & \textbf{0.57} & \textbf{0.43} & \textbf{0.57} & 71.11 \\
    \bottomrule
    \end{tabular}
    }
    \caption{
    \textbf{QG and QA performance of VLMs trained by our framework}. 
    Since we target visual question generation, this table mainly focuses on reporting fine-grained QG scores (normalized to $[0,1]$) while providing average QA accuracy (on five benchmarks) to measure our method's effects on QA performance. 
    Our method consistently improves QG quality across all evaluated backbones, while maintaining competitive QA performance.
    }
    \label{tab:main_results}
\end{table*}

%% file: Table/ablation_dataformat_v3.tex
\begin{table*}[t]
    \centering
    \small
    \setlength{\tabcolsep}{4pt}
    \resizebox{0.8\textwidth}{!}{
    \begin{tabular}{l l ccccc c c}
        \toprule
        \multirow{2}{*}{\textbf{Backbone}} &
        \multirow{2}{*}{\textbf{Setting}} &
        \multicolumn{5}{c}{\textbf{QG Dimension Scores}} &
        \multirow{2}{*}{\textbf{QG Avg.}} &
        \multirow{2}{*}{\textbf{QA Avg.}} \\
        \cmidrule(lr){3-7}
        & &
        \textbf{Search} &
        \textbf{Coverage} &
        \textbf{Context} &
        \textbf{Spatial} &
        \textbf{Diversity} &
        &
        \\
        \midrule
        \multirow{3}{*}{Qwen2.5-VL-3B}
        & QA-only
        & 0.37 & 0.48 & 0.37 & 0.25 & \textbf{0.37} & 0.37 & \textbf{62.40} \\
        & QG-only
        & 0.39 & 0.53 & 0.43 & 0.35 & 0.35 & 0.41 & 60.52 \\
        & QA+QG
        & \textbf{0.43} & \textbf{0.55} & \textbf{0.55} & \textbf{0.37} & 0.35 & \textbf{0.45} & 62.26 \\
        \midrule
                \multirow{3}{*}{Qwen2.5-VL-7B}
        & QA-only
        & 0.45 & 0.48 & 0.38 & 0.28 & 0.35 & 0.39 & 66.57 \\
        & QG-only
        & \textbf{0.48} & \textbf{0.57} & 0.56 & 0.33 & 0.36 & 0.46 & 65.83 \\
        & QA+QG
        & 0.46 & \textbf{0.57} & \textbf{0.59} & \textbf{0.35} & \textbf{0.38} & \textbf{0.47} & \textbf{67.02} \\
        \midrule
        \multirow{3}{*}{Qwen3VL-4B}
        & QA-only
        & 0.47 & 0.57 & 0.47 & 0.37 & 0.33 & 0.44 & 70.38 \\
        & QG-only
        & 0.51 & 0.61 & 0.59 & \textbf{0.54} & 0.35 & 0.52 & 70.12 \\
        & QA+QG
        & \textbf{0.52} & \textbf{0.63} & \textbf{0.62} & \textbf{0.54} & \textbf{0.38} & \textbf{0.54} & \textbf{71.12} \\
        \bottomrule
    \end{tabular}
    }
    \caption{
    \textbf{Effect of supervision format on QG quality and QA performance}.
    We compare QA-only, QG-only, and joint QA+QG supervision under matched settings.
    QG-only training improves question-generation quality over QA-only training, but can reduce average QA performance.
    In contrast, joint QA+QG supervision generally achieves strong QG dimension scores while maintaining or improving average QA accuracy.
    These results suggest that QG-format supervision is important for improving visual questioning capability, while QA-format supervision helps preserve answering performance during adaptation.    
    }
    \label{tab:ablation_supervision}
\end{table*}

%% file: Table/ablation_generationprocess_v2.tex
\begin{table*}[t]
    \centering
    \small
    \setlength{\tabcolsep}{4.5pt}
    \resizebox{0.75\textwidth}{!}{
    \begin{tabular}{c c c c c c c c}
        \toprule
        \multicolumn{2}{c}{\textbf{Generation Process}} &
        \multicolumn{5}{c}{\textbf{QG Dimension Scores}} &
        \multirow{2}{*}{\textbf{QG Avg.}} \\
        \cmidrule(lr){1-2}
        \cmidrule(lr){3-7}
        \textbf{Rewriting} &
        \textbf{Filtering} &
        \textbf{Search} &
        \textbf{Coverage} &
        \textbf{Context} &
        \textbf{Spatial} &
        \textbf{Diversity} &
        \\
        \midrule
        -- & --
        & 0.26 & 0.39 & 0.30 & 0.03 & 0.28 & 0.25 \\
        \xmark & \xmark
        & 0.31 & 0.50 & 0.47 & 0.20 & 0.26 & 0.35 \\
        \cmark & \xmark
        & 0.35 & 0.52 & 0.51 & 0.29 & 0.32 & 0.40 \\
        \cmark & \cmark
        & \textbf{0.43} & \textbf{0.55} & \textbf{0.55} & \textbf{0.37} & \textbf{0.35} & \textbf{0.45} \\
        \bottomrule
    \end{tabular}
    }
    \caption{
    \textbf{Effect of rewriting and filtering procedures in the generation process}. 
    All generated-data variants improve QG quality over the base model. 
    Adding rewriting improves all QG dimensions, and the full pipeline with both rewriting and filtering achieves the strongest overall QG performance, especially on reasoning-related dimensions. 
    QG Avg. is computed as the average over Search, Coverage, Context, Spatial, and Diversity. \looseness-1
    }
    \label{tab:ablation_filtering}
\end{table*}

%% file: Table/ablation_datasource_v2.tex
\begin{table*}[t]
    \centering
    \small
    \setlength{\tabcolsep}{4.0pt}
    \resizebox{0.8\textwidth}{!}{
    \begin{tabular}{l c c c c c c c}
        \toprule
        \multirow{2}{*}{\textbf{Data Source}} &
        \multicolumn{5}{c}{\textbf{QG Dimension Scores}} &
        \multirow{2}{*}{\textbf{QG Avg.}} &
        \multirow{2}{*}{\textbf{QA Avg.}} \\
        \cmidrule(lr){2-6}
        &
        \textbf{Search} &
        \textbf{Coverage} &
        \textbf{Context} &
        \textbf{Spatial} &
        \textbf{Diversity} &
        &
        \\
        \midrule
        Base
        & 0.26 & 0.39 & 0.30 & 0.03 & 0.28
        & 0.25 & 60.49 \\
        Original SAT 
        & \textbf{0.45} & 0.49 & 0.21 & 0.22 & 0.17
        & 0.31 & 57.46 \\
        Original SAT + Self-Supervision
        & 0.44 & 0.54 & 0.54 & 0.32 & 0.34
        & 0.44 & 61.28 \\
        Self-Supervision
        & 0.43 & \textbf{0.55} & \textbf{0.55} & \textbf{0.37} & \textbf{0.35}
        & \textbf{0.45} & \textbf{62.26} \\
        \bottomrule
    \end{tabular}
    }
    \caption{
    \textbf{Effect of self-supervision data on training}. 
    Self-supervision data refers to the question-answer supervision generated through our rewriting and filtering process. 
    All variants use the same SAT image source and QA+QG training format, but differ in the source of supervision. 
    Original SAT supervision improves visual search difficulty but degrades QA performance and performs worse on reasoning and diversity dimensions. 
    In contrast, our self-supervision data achieves the strongest overall QG profile and the best QA average.
    QG Avg. is computed as the average over Search, Coverage, Context, Spatial, and Diversity.
    }
    \label{tab:ablation_source}
\end{table*}

%% file: Table/ablation_betterquestion.tex
\begin{table}[t]
    \centering
    \small
    \setlength{\tabcolsep}{4pt}
    \resizebox{\columnwidth}{!}{
    \begin{tabular}{lcccccc}
    \toprule
    \textbf{Setting} & \textbf{CVB-2D} & \textbf{CVB-3D} & \textbf{Spar} & \textbf{VStar} & \textbf{RWQA} & \textbf{Avg.} \\
    \midrule
    Base-Q      
    & 69.16 & 69.25 & \textbf{30.21} & 75.92 & 64.97 & 61.90 \\
    Improved-Q  
    & \textbf{69.19} & \textbf{75.58} & 29.78 & \textbf{76.44} & \textbf{65.62} & \textbf{63.32} \\
    \bottomrule
    \end{tabular}
    }
    \caption{
        \textbf{Effect of generated question quality on QA training}. 
        Both settings use GPT-generated answers, but differ in whether the training questions are generated by the base model (\textbf{Base-Q}) or the improved QG model (\textbf{Improved-Q}). 
        Improved-Q achieves stronger average QA performance, suggesting that higher-quality generated questions provide more useful supervision.
    }
    \label{tab:ablation_gpt_answer}
\end{table}

%% file: Section/related_work.tex
\section{Related Work}
\label{sec:related_work}

\subsection{Visual Question Generation}

Visual Question Generation (VQG) aims to generate natural-language questions grounded in image content. 
Existing VQG methods are commonly built from external supervision, including VQA-style data\citep{mi2024convqg,shen2024ask,vedd2022guiding,zhao2024lova3}, LLM-generated questions\citep {suwono2023location,zhang2025diagram}, and controllable VQG data with predefined constraints such as answers, regions, knowledge, or difficulty levels\citep{fang2024diverse,mi2024convqg,uehara2023k,shen2024ask,xie2025explicitly2,xie2025explicitly}. 
Although these methods improve answerability, grounding, diversity, or difficulty control, their question distributions remain tied to fixed datasets, external generators, or predefined control spaces, limiting the autonomous development of stronger visual questioning capability.

VQG evaluation is also often tied to reference-based metrics such as BLEU, METEOR, CIDEr to compare similarity to given ground-truth questions\citep{mi2024convqg,vedd2022guiding,xie2025explicitly2,xie2025explicitly}. 
Recent work uses LLMs or MLLMs as judges for multimodal outputs or generated questions\citep{chen2024mllm,noorbakhsh2025savaal,yao2025mcqg}, but mainly evaluates general multimodal response quality or text-domain question quality rather than fine-grained image-grounded questioning capability. 

\subsection{Post-Training for VLMs}

Post-training improves language and vision-language models with instruction data collected from humans, curated datasets, or stronger models\citep{liu2023visual,wang2024vigc,zhou2026learning}. However, constructing high-quality supervision is costly and difficult to scale. 
Recent self-improvement methods reduce this dependence by letting models generate, filter, critique, or judge their own training signals\citep{yuan2024self,shen2025cycle,deng2024enhancing,cheng2025vision}. 
However, these works mainly use self-generated supervision to improve answering-oriented capabilities, such as visual instruction following, reasoning, and QA performance\citep{wang2024vigc,deng2024enhancing,cheng2025vision,pan2026through}. 
Unlike these works, where generated questions mainly serve as intermediate supervision or interaction signals, our work makes the visual questioner itself the object of post-training while preserving answering ability.



%% file: Section/conclusion.tex
\section{Conclusion}
\label{sec:conclusion}

We present a self-evolving visual questioner framework that improves a VLM's QG capability without human annotations or external teacher models. Through question proposal, rewriting, filtering, and dual-format QA/QG training, our method constructs visual-centric, diverse, and reasoning-oriented supervision from unlabeled images. We further introduce an evaluation protocol covering perception, reasoning, and diversity. Experiments show substantial QG improvements while largely preserving QA performance, and ablations validate the importance of each component and the downstream value of improved questions.

\section{Limitations}
\label{sec:exp_limitations}
Despite the effectiveness of our self-evolving framework, it still has several limitations that could be addressed in future work. 
First, our filtering process currently focuses on answerability and perception/reasoning difficulty, while more fine-grained criteria could further capture visual grounding, ambiguity, and instructional value. 
Second, the proposal-rewriting-filtering-training loop introduces additional computation compared with directly training on existing QA data, especially when applied iteratively. 
This overhead could be reduced through batching, lightweight filtering, or more selective evolution strategies. 
Nevertheless, the observed improvements in question quality suggest that self-evolving visual questioners are a promising direction for building stronger and more scalable questioning-oriented VLMs.

%% file: Section/Appendix/detailed_qg_evaluation.tex
\section{Detailed Explanation of Questioning Capability Evaluation}
\label{app:detail_QG_definition}

Our evaluation protocol measures whether generated questions are visually grounded, challenging, and non-redundant. 
As summarized in Table~\ref{tab:qg_eval_dimensions}, we evaluate generated questions at two levels. 
At the individual-question level, we measure four dimensions grouped into \textbf{perception difficulty} and \textbf{reasoning difficulty}. 
Perception difficulty captures what visual evidence must be found, while reasoning difficulty captures how that evidence must be interpreted. 
At the question-set level, we measure \textbf{questioning diversity}, which captures whether questions generated for the same image provide complementary rather than redundant supervision.

\begin{table*}[t]
    \centering
    \small
    \setlength{\tabcolsep}{4pt}
    \begin{tabular}{l l p{0.62\linewidth}}
        \toprule
        \textbf{Level} & \textbf{Dimension} & \textbf{What it measures} \\
        \midrule
        Question & \textbf{Visual Search Difficulty} 
        & How difficult it is to locate the visual evidence needed to answer the question. \\
        Question & \textbf{Visual Evidence Coverage} 
        & How broadly the question depends on evidence across objects, regions, attributes, or scene cues. \\
        Question & \textbf{Visual Context Reasoning} 
        & Whether the question requires contextual interpretation of visible cues beyond direct recognition. \\
        Question & \textbf{Visual Spatial Reasoning} 
        & Whether the question requires reasoning over spatial relations, layout, orientation, or scene structure. \\
        Set & \textbf{Questioning Diversity} 
        & How non-redundant the questions generated for the same image are. \\
        \bottomrule
    \end{tabular}
    \caption{Summary of our visual questioning capability evaluation dimensions.}
    \label{tab:qg_eval_dimensions}
\end{table*}

\subsection{Perception Difficulty}

Perception difficulty evaluates what visual evidence must be observed to answer a question. 
It captures the demand placed on visual inspection before reasoning: whether the answer depends on salient evidence, subtle details, localized regions, or broader image content. 
We decompose perception difficulty into two dimensions: \textit{Visual Search Difficulty} and \textit{Visual Evidence Coverage}.

\paragraph{Visual Search Difficulty.}
\textit{Visual Search Difficulty} measures how difficult it is to locate and inspect the evidence needed to answer a question. 
Questions that depend on salient objects, obvious colors, or clearly visible main subjects receive lower scores. 
Questions receive higher scores when answering them requires careful inspection of subtle evidence, small or partially occluded objects, peripheral regions, fine-grained details, relative size or area, or counting.

\paragraph{Visual Evidence Coverage.}
\textit{Visual Evidence Coverage} measures how broadly a question depends on evidence across the image. 
Questions centered on a single dominant object or a single simple attribute receive lower scores. 
Questions receive higher scores when they require evidence from multiple objects, regions, attributes, scene cues, or the overall arrangement of the image. 
This dimension is independent of search difficulty: a question may rely on broad image evidence even if each individual cue is easy to locate.

\subsection{Reasoning Difficulty}

Reasoning difficulty evaluates how the observed visual evidence must be used to answer a question. 
While perception difficulty focuses on finding and inspecting evidence, reasoning difficulty focuses on the inference required after the relevant evidence has been identified. 
We decompose reasoning difficulty into two dimensions: \textit{Visual Context Reasoning} and \textit{Visual Spatial Reasoning}.

\paragraph{Visual Context Reasoning.}
\textit{Visual Context Reasoning} measures whether a question requires interpreting what visible cues imply beyond direct recognition or description. 
Questions that can be answered by simply naming objects, reading visible attributes, or describing directly observable content receive lower scores. 
Questions receive higher scores when they require inferring the scene context, object state, activity, function, intention, cause, or likely situation suggested by the visual evidence.

\paragraph{Visual Spatial Reasoning.}
\textit{Visual Spatial Reasoning} measures whether a question requires reasoning about spatial relationships among visible elements. 
Questions that do not depend on spatial structure receive lower scores. 
Questions receive higher scores when they require comparing positions, orientations, distances, containment, occlusion, accessibility, layout, object interactions, or multi-hop spatial relations within the scene.

\subsection{Questioning Diversity}

\textit{Questioning Diversity} is measured at the question-set level because diversity cannot be determined from a single question in isolation. 
For each image $x$, we consider the generated question set $\mathcal{Q}_x=\{q_1,\ldots,q_m\}$. 
We sample $n$ questions from $\mathcal{Q}_x$ and repeat this sampling over multiple runs. 
Each sampled question is encoded by a sentence embedding model and $\ell_2$-normalized:
\begin{equation}
    z_i = \frac{f(q_i)}{\|f(q_i)\|_2}.
\end{equation}
We then compute the average pairwise cosine similarity within the sampled set:
\begin{equation}
    s_{\mathrm{emb}}(\mathcal{Q}_x) = \frac{2}{n(n-1)} \sum_{1 \leq i < j \leq n} z_i^\top z_j.
\end{equation}
The diversity score is defined as the corresponding average pairwise cosine distance:
\begin{equation}
    d_{\mathrm{div}}(\mathcal{Q}_x) = 1 - s_{\mathrm{emb}}(\mathcal{Q}_x).
\end{equation}

Higher values indicate that the questions generated for the same image are semantically less redundant and cover more distinct aspects of the image. 
Lower values indicate repeated templates, near-paraphrases, or duplicated answer targets. 
This set-level metric complements the individual-question rubric scores by measuring whether the generated supervision is diverse across questions, rather than only strong for each question independently.

%% file: Section/Appendix/Prompts.tex
\section{Detailed Prompts}
\label{app:prompts}

In this section, we provide the detailed prompts used for our data generation process and evaluation process. Table~\ref{tab:prompt_overview} summarizes the prompts used in our framework. The full prompt texts are shown in Fig.~\ref{fig:prompt_seed_qg}--Fig.~\ref{fig:prompt_qg_eval}.

\begin{table}[h]
\centering
\small
\begin{tabular}{l l}
\toprule
\textbf{Stage} & \textbf{Reference} \\
\midrule
Question Proposal & Fig.~\ref{fig:prompt_seed_qg} \\
Rewriting with visual facts & Fig.~\ref{fig:prompt_self_evolved_broader} \\
Rewriting with hard examples & Fig.~\ref{fig:prompt_self_evolved_harder} \\
Filtering & Fig.~\ref{fig:prompt_self_filter_part1} \& \ref{fig:prompt_self_filter_part2} \\
Agentic QG evaluation & Fig.~\ref{fig:prompt_qg_eval} \\
\bottomrule
\end{tabular}
\caption{Overview of prompts used in the self-evolving visual questioner framework and the agentic visual questioning capability evaluation.}
\label{tab:prompt_overview}
\end{table}

\begin{figure*}[t]
\begin{tcolorbox}[colback=gray!5, colframe=black!45,
    fonttitle=\bfseries, title={Question Proposal}]
\small
\lstset{breaklines=true, basicstyle=\footnotesize\ttfamily}

\begin{lstlisting}
List n1 topics that you can answer questions about. Choose a topic uniformly from this list, and state it. Then write 60 subtopics about the chosen topic. Then choose a subtopic uniformly from this list, and state it. Then write a question that is not about the subtopic, but can only be answered with expertise in the subtopic. Then write the answer. Both the question and answer should be long. The name of the subtopic should not appear in the question, and none of the words in the subtopic should be reused in the question. 
The question must rely on visual information such that its answer would change or be impossible to determine without the image. The answer should be deterministic and answerable with a few sentences. You should generate only ONE focused question, not multiple sub-questions. IMPORTANT: Do NOT create compound or multi-part questions by joining sub-questions with 'and' or 'what'. A question like 'How does X affect Y, and what in the image shows Z?' is FORBIDDEN. It is two easy questions concatenated, not a harder one. Instead, make the SINGLE question harder by requiring deeper reasoning: multi-hop inference, counterfactual/causal reasoning, cross-region comparison, or quantitative/ordinal judgment. The answer must be DIFFERENT from the original answer. 
Return the result in JSON format: {'Question': 'xxx', 'Thinkings': 'Your selection process'}.
\end{lstlisting}

\end{tcolorbox}
\caption{Question proposal prompt used to produce the initial question-answer pair.}
\label{fig:prompt_seed_qg}
\end{figure*}

\begin{figure*}[t]
\begin{tcolorbox}[colback=gray!5, colframe=black!45,
    fonttitle=\bfseries, title={Rewriting Prompt for Broader Questions}]
\small
\lstset{breaklines=true, basicstyle=\footnotesize\ttfamily}

\begin{lstlisting}
You are a Question generator. Your objective is to rewrite a given Q&A into a genuinely harder version that requires more visual evidence or reasoning steps. Make it harder by increasing the DEPTH of reasoning required. Examples include evidence selection, object-role comparison, spatial relation reasoning, scene-function inference, and multi-object grounding. Do NOT make it harder by simply appending more sub-questions or extra clauses. Do NOT generate a question that asks the same thing as the original or is just a paraphrase. Do NOT start your rewritten question with the same first 5 words as the seed question. The question should not become broader, more subjective, or more generic. The added constraint must change the type or depth of reasoning required, not just expand the question's length.

Use the extracted visual facts below as grounding evidence. Only use facts that are clearly supported by the image.

Extracted visual facts:
{rag_facts}

Question: {seed_question}
Answer: {seed_answer}

Constraints:
- Require at least 2 concrete visible objects, regions, or relations.
- The question must contain a hidden intermediate inference.
- The intermediate inference step must not be given away in the question.
- Do NOT ask for subjective preference or broad commonsense.
- Do NOT merely paraphrase the seed question with synonyms.
- Ensure all generated data is consistent with the image content.
- The question must rely on visual information such that its answer would change or be impossible to determine without the image.
- Generate only ONE focused question, not multiple sub-questions.
- Make the SINGLE question harder by requiring deeper reasoning: multi-hop inference, counterfactual/causal reasoning, cross-region comparison, or quantitative/ordinal judgment.
- 'reasoning_steps' MUST contain at least 2 non-trivial, concrete operations, not just 'step 1' or 'step 2'.

Return the result in JSON format:
{'Rewritten question': 'one single syntactic question ending with ?', 'visible_evidence': ['at least two concrete visible objects/regions/relations required to answer, using the EXACT object names that appear in your question'], 'intermediate_inference': 'the hidden bridging inference required before the final answer', 'reasoning_steps': ['<concrete step, e.g. Locate the red chair>', '<concrete step, e.g. Compare its distance to the desk lamp vs the blue chair>'], 'why_harder_than_seed': 'why this requires more grounded reasoning than the seed',}
\end{lstlisting}

\end{tcolorbox}
\caption{Rewriting prompt that rewrites a question proposal using extracted visual facts as grounding evidence.}
\label{fig:prompt_self_evolved_broader}
\end{figure*}

\begin{figure*}[t]
\begin{tcolorbox}[colback=gray!5, colframe=black!45,
    fonttitle=\bfseries, title={Rewriting Prompt for Harder Questions}]
\small
\lstset{breaklines=true, basicstyle=\footnotesize\ttfamily}

\begin{lstlisting}
You are a Question generator. Your objective is to rewrite a given Q&A into a genuinely harder version that requires more visual evidence or reasoning steps. Make it harder by increasing the DEPTH of reasoning required. Examples include evidence selection, object-role comparison, spatial relation reasoning, scene-function inference, and multi-object grounding. Do NOT make it harder by simply appending more sub-questions or extra clauses. Do NOT generate a question that asks the same thing as the original or is just a paraphrase. Do NOT start your rewritten question with the same first 5 words as the seed question. The question should not become broader, more subjective, or more generic. The added constraint must change the type or depth of reasoning required, not just expand the question's length.

The following questions were generated for similar images and are rated as HARD:
{examples_text}

Rewrite the following Q&A so it is harder than the examples while staying tightly grounded in this image. It must require more reasoning steps, finer perceptual discrimination, or a deeper multi-hop chain.

Question: {seed_question}
Answer: {seed_answer}

Examples of GOOD rewritten questions:
Example 1:
Q: 'Which chair, the wooden chair or the metal chair, is positioned closer to the desk lamp?'
visible_evidence: ['wooden chair', 'metal chair', 'desk lamp']
reasoning_steps: ['Locate the wooden chair and measure its distance to the desk lamp', 'Locate the metal chair and measure its distance to the desk lamp', 'Compare the two distances']

Constraints:
- Do not reuse the same reasoning pattern as any example above.
- The added difficulty must come from deeper reasoning, not longer question text.
- Require at least 2 concrete visible objects, regions, or relations.
- The question must contain a hidden intermediate inference.
- Do NOT merely paraphrase the seed question with synonyms.
- Ensure all generated data is consistent with the image content.
- The question must rely on visual information such that its answer would change or be impossible to determine without the image.
- Generate only ONE focused question, not multiple sub-questions.
- 'reasoning_steps' MUST contain at least 2 non-trivial, concrete operations, not just 'step 1' or 'step 2'.

Return the result in JSON format:
{'Rewritten question': 'one single syntactic question ending with ?', 'visible_evidence': ['at least two concrete visible objects/regions/relations required to answer, using the EXACT object names that appear in your question'], 'intermediate_inference': 'the hidden bridging inference required before the final answer', 'reasoning_steps': ['<concrete step, e.g. Locate the red chair>', '<concrete step, e.g. Compare its distance to the desk lamp vs the blue chair>'], 'why_harder_than_seed': 'why this requires more grounded reasoning than the seed', 'final_answer_type': 'short phrase|entity|comparison|count|yes_no', }
\end{lstlisting}

\end{tcolorbox}
\caption{Rewriting prompt that rewrites a question proposal using retrieved hard examples as references.}
\label{fig:prompt_self_evolved_harder}
\end{figure*}

\begin{figure*}[t]
\begin{tcolorbox}[colback=gray!5, colframe=black!45,
    fonttitle=\bfseries, title={Filtering Prompt (Part 1)}]
\small
\lstset{breaklines=true, basicstyle=\footnotesize\ttfamily}

\begin{lstlisting}[breaklines=true, basicstyle=\ttfamily\small]
You are a strict filtering judge for visual question generation.

Your task is to evaluate whether a generated question-answer pair should be kept for training. You must judge the question using the provided image, the seed question-answer pair, and the generated question-answer pair.

You must perform four checks:
1. Entity extraction: extract all noun or noun-phrase entities mentioned in the generated question.
2. Visual entity difficulty: categorize each shared visual entity as EASY or HARD based on the image.
3. Reasoning difficulty: infer the cognitive reasoning steps required to answer the generated question.
4. filter decision: decide whether the generated question is better than the seed question and should be kept.

Definitions:
- EASY visual entity: prominent, unobstructed, and immediately identifiable in the image.
- HARD visual entity: tiny, occluded, blurry, low-contrast, visually similar to nearby objects, peripheral, or requires careful inspection.
- Reasoning step: one cognitive operation required to answer the question, such as identifying objects, comparing quantities, interpreting spatial relations, inferring function, integrating multiple cues, or reasoning about cause/effect.
- Do not count meta steps such as "look at the image" or "analyze the question".
- Do not include verification or double-checking steps.
- Do not include the final answer as a reasoning step.

Keep the generated question only if ALL conditions below are satisfied:
- It is answerable from the image.
- Its answer is consistent with the image.
- It is not a trivial rephrasing of the seed question.
- It is visually grounded and depends on image evidence.
- It is at least as difficult as the seed question in one or more of the following dimensions:
  visual perception difficulty, text perception difficulty, or reasoning difficulty.
- It does not become vague, subjective, overly broad, or multi-question.
- It asks one focused question, not several questions joined together.
- Its answer is deterministic and concise.
\end{lstlisting}

\end{tcolorbox}
\caption{(Part 1) Filtering prompt used to determine whether a rewritten question should be retained in the self-evolving visual questioner framework..}
\label{fig:prompt_self_filter_part1}
\end{figure*}

\begin{figure*}[t]
\begin{tcolorbox}[colback=gray!5, colframe=black!45,
    fonttitle=\bfseries, title={Filtering Prompt (Part 2)}]
\small
\lstset{breaklines=true, basicstyle=\footnotesize\ttfamily}

\begin{lstlisting}[breaklines=true, basicstyle=\ttfamily\small]
Inputs:
Image: [VISUAL_INPUT]

Seed question:
{seed_question}

Seed answer:
{seed_answer}

Generated question:
{generated_question}

Generated answer:
{generated_answer}

Output a SINGLE minified JSON object only.
Do not output markdown, explanations, comments, or extra text.
The JSON must start with { and end with }.

Required JSON schema:
{
  "entity_extraction": {
    "generated_question_entities": ["entity1", "entity2"]
  },
  "visual_entity_difficulty": {
    "easy_count": 0,
    "hard_count": 0,
    "categorized_entities": [
      {
        "entity": "name",
        "difficulty": "easy|hard",
        "reason": "short reason"
      }
    ]
  },
  "reasoning_difficulty": {
    "count": 0,
    "evidence": [
      "step1",
      "step2"
    ]
  }
}

Important rules:
- Analyze every extracted entity that is visible or potentially grounded in the image.
- The sum of easy_count and hard_count must equal the number of unique visual entities analyzed.
- If the generated question cannot be answered from the image, set decision to "discard".
- If the generated answer contradicts the image, set decision to "discard".
- If the question contains multiple sub-questions, set single_focused_question to false and decision to "discard".
- If none of improves_visual_perception, improves_text_perception, or improves_reasoning is true, set decision to "discard".
\end{lstlisting}

\end{tcolorbox}
\caption{(Part 2) Filtering prompt used to determine whether a rewritten question should be retained in the self-evolving visual questioner framework..}
\label{fig:prompt_self_filter_part2}
\end{figure*}


\begin{figure*}[t]
\begin{tcolorbox}[colback=gray!5, colframe=black!45,
    fonttitle=\bfseries, title={Question Quality Evaluation Prompt}]
\small
\lstset{breaklines=true, basicstyle=\footnotesize\ttfamily}

\begin{lstlisting}
Use only the image and the current question. Ignore model identity and do not compare with other questions. Score what the question requires to answer, not how long or fluent it is. Give credit to visually grounded conceptual or functional questions when the image contains concrete evidence; penalize generic world-knowledge answers, unsupported assumptions, and unanswerable questions.

Record:
{record_json}

Score independently from 0 to 5:
- visual_search_difficulty: visual search/inspection effort. 0=no image needed; 1=obvious evidence; 3=specific region or nearby items; 5=multi-region search, counting, distance/area, or easy-to-miss fine detail.
- visual_evidence_coverage: breadth of concrete visual evidence. 0=no visible evidence; 1=one dominant object/scene; 3=two objects or one local region; 5=broad scene evidence, multiple important regions, or overall arrangement.
- visual_context_reasoning: conceptual interpretation beyond direct observation. 0=non-visual/unsupported; 1=direct identifying, reading, counting, locating, or comparing; 2=simple visible state/action/category judgment; 3=single-step interpretation from one cue; 4=multi-cue scene/situation interpretation; 5=non-obvious explanation or implication requiring weighed visual cues.
- visual_spatial_reasoning: use of layout/position/object relations. 0=none; 1=absolute location; 2=one simple relation; 3=one spatial comparison; 4=multiple linked relations; 5=complex spatial inference such as path, occlusion, containment, stability, or scene structure.

Return JSON only:
{
  "visual_search_difficulty": int,
  "visual_evidence_coverage": int,
  "visual_context_reasoning": int,
  "visual_spatial_reasoning": int
}
\end{lstlisting}

\end{tcolorbox}
\caption{Question quality evaluation prompt used to score generated questions along visual search, evidence coverage, contextual reasoning, and spatial reasoning dimensions.}
\label{fig:prompt_qg_eval}
\end{figure*}

%% file: Section/Appendix/Training_parameter.tex
\section{Implementation Details}
\label{app:training_parameters}

\paragraph{Training setup.}
All models are trained with standard supervised fine-tuning from their corresponding instruction-tuned checkpoints. 
We conduct experiments with Qwen2.5-VL-3B-Instruct, Qwen2.5-VL-7B-Instruct, and Qwen3VL-4B-Instruct. 
Unless otherwise specified, each model is trained on $10{,}000$ retained question-answer pairs produced by our self-evolution pipeline. 
We use a training batch size of $2$ and AdamW as the optimizer. 
The learning rate is set to $2 \times 10^{-5}$ with a cosine scheduler and a warm-up ratio of $0.1$. 
All models are trained for one epoch.

\paragraph{Computing infrastructure and budget.}
Training and evaluation for Qwen2.5-VL-3B-Instruct and Qwen3-VL-4B-Instruct are conducted on NVIDIA RTX A5000 GPUs, while Qwen2.5-VL-7B-Instruct is trained and evaluated on NVIDIA A100 GPUs. 
Most training, generation, and evaluation runs use a single GPU per run. 
For data generation, producing one training-ready set of $10{,}000$ self-supervision question-answer pairs requires approximately 24 GPU hours with 16-way parallel generation. 
For supervised fine-tuning, training one model on $10{,}000$ retained examples for one epoch takes approximately $1-3$ GPU hours, depending on the model size. 
For evaluation, running the full benchmark suite for one checkpoint takes approximately $4-7$ GPU hours, depending on the model size. 
Across the main experiments, ablations, and evaluation runs, the total computational budget is estimated from these per-run costs and includes both successful runs and exploratory runs used to tune training settings. 
All experiments are run on a Linux GPU cluster with standard CUDA/PyTorch-based training infrastructure.

\paragraph{Evaluation setup.}
For question-generation evaluation, all compared models use the same held-out image set, generation prompt, and decoding configuration. 
We use GPT-5.4 as the judge model for individual-question evaluation, including visual search difficulty, visual evidence coverage, visual context reasoning, and visual spatial reasoning. 
For set-level diversity evaluation, we use \texttt{Qwen/Qwen3-Embedding-4B} as the embedding model and compute semantic distances among questions generated for the same image. 
For QG evaluation, we compute scores by averaging evaluation results across generated questions or images, depending on the evaluation dimension. For downstream QA evaluation, we follow the standard protocol of each benchmark, keep inference settings fixed across all compared checkpoints, and report accuracy averaged over three runs.

\paragraph{Software and evaluation implementations.}
For model training and question generation, we use the official Qwen2.5-VL and Qwen3-VL implementations through the standard PyTorch and Hugging Face Transformers package. 
For QG evaluation, we use GPT-5.4 as the judge model with fixed prompts, rubrics, and decoding settings across all compared methods. 
For diversity evaluation, we use \texttt{Qwen/Qwen3-Embedding-4B} as the embedding model and compute semantic distances among questions generated for the same image. 
For downstream QA evaluation, we use VLMEvalKit to evaluate benchmark performance and follow the standard evaluation protocols and answer-matching rules of each benchmark. 
We do not modify the underlying model architectures or external library implementations.

%% file: Section/Appendix/detailed_qa_performance.tex
\section{Detailed QA Benchmark Results}
\label{app:qa_details}

\input{Table/detailed_qa}

Table~\ref{tab:appendix_qa_details} provides the full QA benchmark results corresponding to the main experiments and the data-source ablation. 
The evaluation includes CVBench-2D with 1,438 samples, CVBench-3D with 1,200 samples, SPARBench with 7,207 samples, VStar with 191 samples, and RealWorldQA with 765 samples. 
These results support the conclusion that our filtered self-generated data improves question-generation quality without causing broad degradation in downstream answering ability.

%% file: Table/detailed_qa.tex
\begin{table*}[t]
    \centering
    \small
    \setlength{\tabcolsep}{4.0pt}
    \resizebox{0.8\textwidth}{!}{
    \begin{tabular}{l l c c c c c c}
        \toprule
        \textbf{Backbone} &
        \textbf{Training stage / Data source} &
        \textbf{CVB-2D} &
        \textbf{CVB-3D} &
        \textbf{Spar} &
        \textbf{VStar} &
        \textbf{RWQA} &
        \textbf{Avg.} \\
        \midrule
        \multirow{8}{*}{Qwen2.5-VL-3B}
        & Base
        & 66.75 & 65.50 & 28.89 & 74.39 & 66.92 & 60.49 \\
        & First-round
        & 69.35 & 71.50 & 30.50 & 74.87 & 65.10 & 62.26 \\
        & Second-round
        & 69.31 & 72.17 & 31.24 & 75.39 & 64.70 & 62.56 \\
        \cmidrule(lr){2-8}
        & QA-only
        & 69.28 & 71.67 & 30.96 & 74.87 & 65.23 & 62.40 \\
        & QA+QG
        & 69.34 & 71.50 & 30.50 & 74.87 & 65.10 & 62.26 \\
        \cmidrule(lr){2-8}
        & Original SAT
        & 64.87 & 61.75 & 31.06 & 68.59 & 61.05 & 57.46 \\
        & Original SAT + Self-Evolving
        & 68.74 & 69.67 & 30.92 & 73.30 & 63.79 & 61.28 \\
        & Self-Evolving
        & 69.34 & 71.50 & 30.50 & 74.87 & 65.10 & 62.26 \\
        \midrule
        \multirow{5}{*}{Qwen2.5-VL-7B}
        & Base
        & 75.67 & 74.83 & 33.81 & 76.44 & 68.76 & 65.90 \\
        & First-round
        & 75.98 & 78.08 & 33.50 & 77.49 & 70.07 & 67.02 \\
        & Second-round
        & 76.29 & 78.42 & 33.55 & 76.96 & 69.28 & 66.90 \\
        \cmidrule(lr){2-8}
        & QA-only
        & 75.19 & 77.67 & 33.61 & 77.49 & 68.89 & 66.57 \\
        & QA+QG
        & 75.98 & 78.08 & 33.47 & 77.49 & 70.07 & 67.02 \\
        \midrule
        \multirow{5}{*}{Qwen3VL-4B}
        & Base
        & 78.83 & 91.92 & 36.16 & 78.53 & 71.11 & 71.31 \\
        & First-round
        & 78.06 & 91.08 & 36.29 & 79.06 & 71.11 & 71.12 \\
        & Second-round
        & 78.39 & 91.50 & 36.16 & 78.53 & 70.98 & 71.11 \\
        \cmidrule(lr){2-8}
        & QA-only
        & 77.39 & 90.83 & 35.71 & 77.49 & 70.46 & 70.38 \\
        & QA+QG
        & 78.06 & 91.08 & 36.29 & 79.06 & 71.11 & 71.12 \\
        \bottomrule
    \end{tabular}
    }
    \caption{
    Detailed QA benchmark results used in the main experiments, supervision-format ablation, and data-source ablation. 
    We report accuracy on CVBench-2D, CVBench-3D, SparBench, VStar, and RealWorldQA, together with the average accuracy across these five benchmarks. 
    The First-round rows correspond to the default self-evolving QA+QG setting used in the main results. 
    The QA-only and QA+QG rows compare supervision formats under matched settings, while the original SAT, original SAT + self-evolving, and self-evolving rows compare different sources of question-answer supervision in the Qwen2.5-VL-3B setting.
    }
    \label{tab:appendix_qa_details}
\end{table*}

%% file: Section/Appendix/human_alignment.tex
\section{Human and QG Evaluation Alignment}
\label{app:human_alignment}

\subsection{Human Annotation Protocol}
\label{app:human_annotation_protocol}

To validate the reliability of our GPT-based question-generation evaluation, we conduct a human-GPT alignment study on a randomly sampled set of 100 generated visual questions. 
The annotators were recruited internally from student researchers familiar with vision-language tasks.
Participation was voluntary and unpaid, and no private or sensitive information was collected from the annotators.
The annotation task only involves scoring generated questions according to the same rubrics used by our automatic evaluator.

\begin{figure*}[t]
    \centering
    \includegraphics[width=0.48\textwidth]{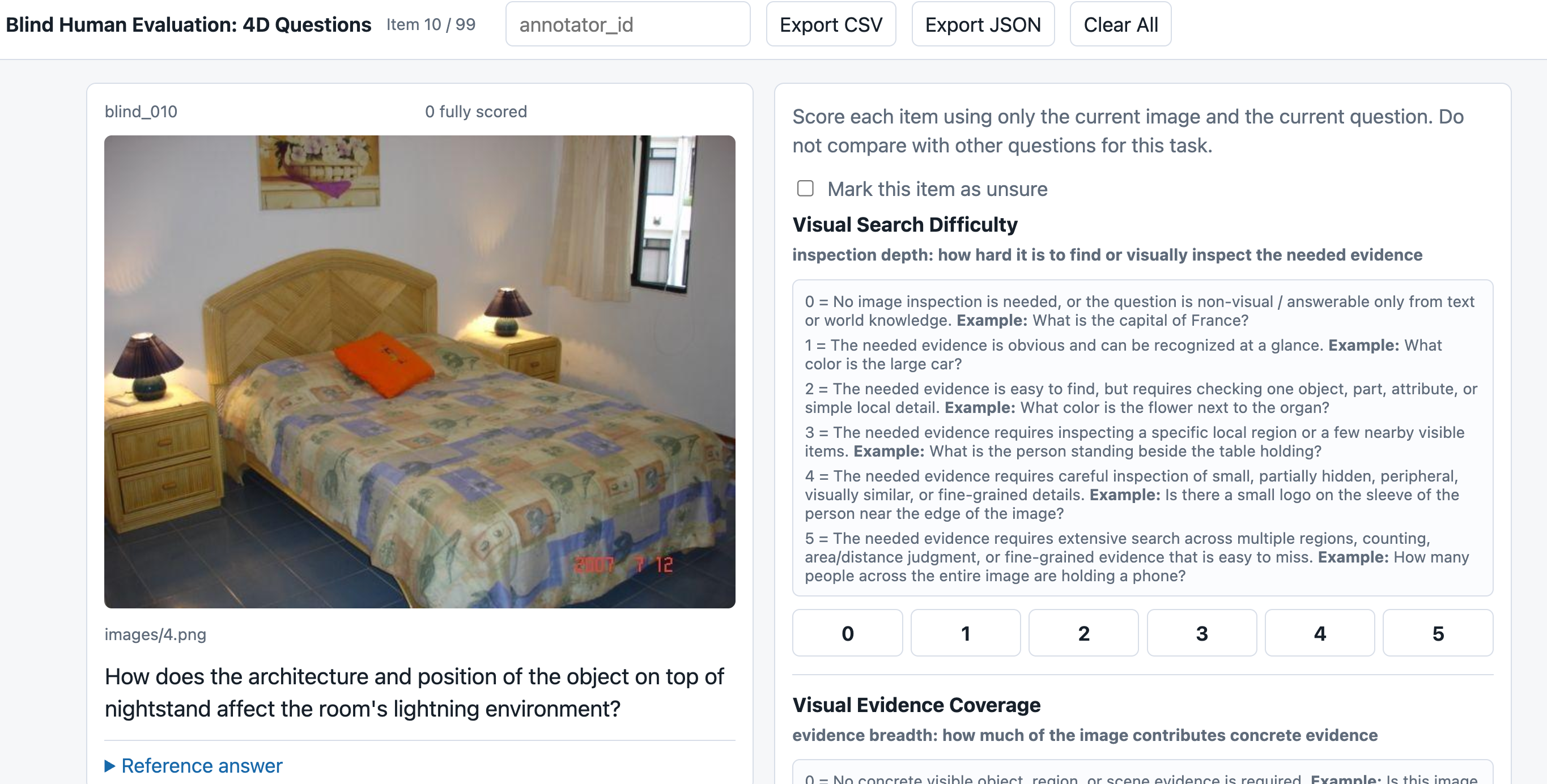}
    \includegraphics[width=0.48\textwidth]{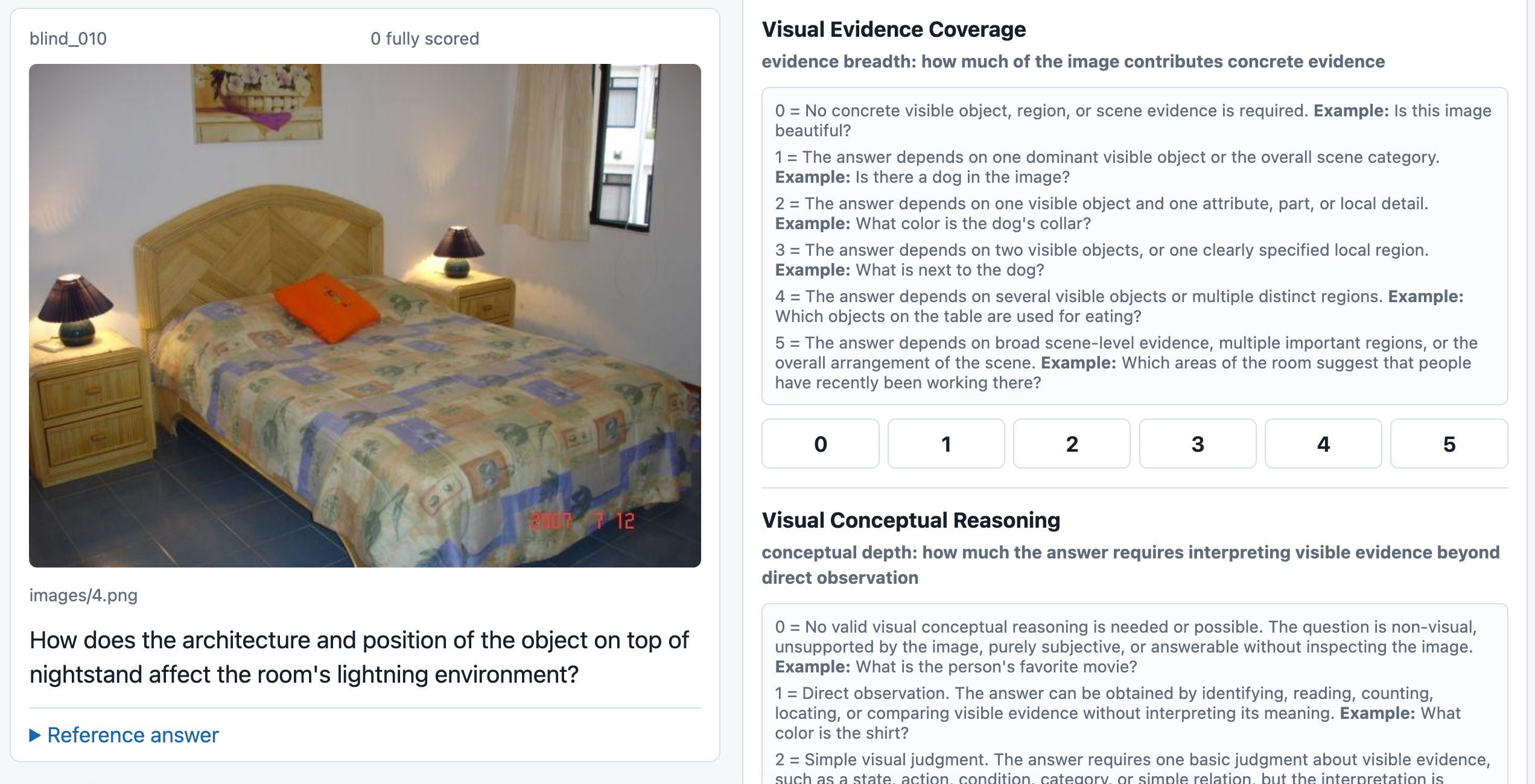}
    
    \vspace{0.5em}
    
    \includegraphics[width=0.48\textwidth]{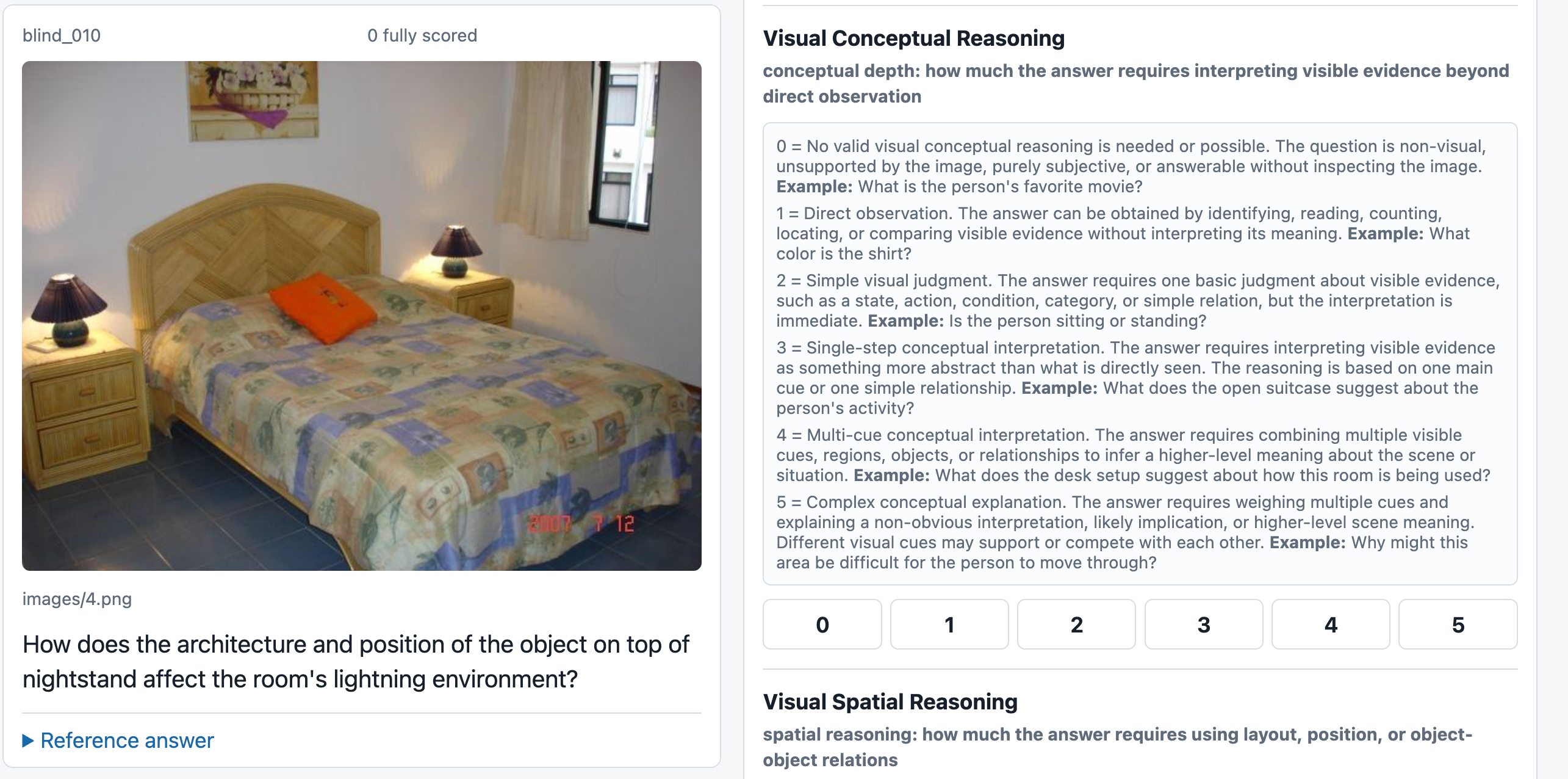}
    \includegraphics[width=0.48\textwidth]{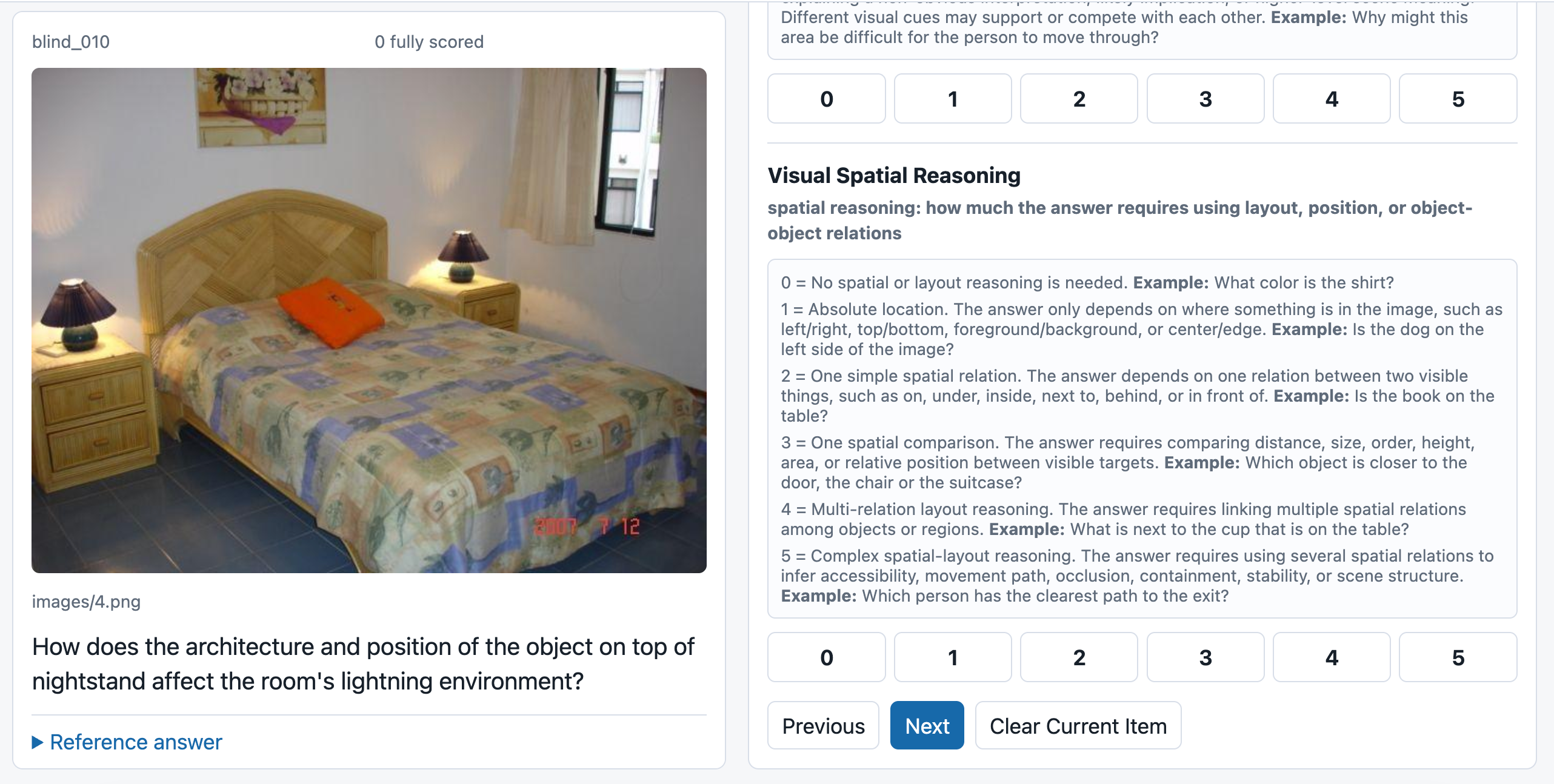}
    \caption{
    Screenshots of the human annotation interface and scoring instructions used in the human--GPT alignment study.
    Annotators are shown the image, generated question, and rubrics for scoring visual search difficulty, visual evidence coverage, visual context reasoning, and visual spatial reasoning. Zoom-in for more details.
    }
    \label{fig:human_screen}
\end{figure*}

Annotators are shown the image, the generated question, and the scoring instructions for each dimension. 
They are asked to independently score four QG dimensions: visual search difficulty, visual evidence coverage, visual context reasoning, and visual spatial reasoning. 
The full annotation interface and instruction screenshots are shown in Figure~\ref{fig:human_screen}. 

The task does not ask annotators to provide personal information, identify individuals in the images, infer protected attributes, or make decisions about people. 
Annotators are instructed to focus only on whether the question is grounded in the image and how much visual perception or reasoning is required to answer it. 
Participation is voluntary, and no private or sensitive information is collected from the annotators. 
The sampled questions are generated from existing public research datasets and are screened to avoid clearly offensive or unsafe content before annotation. 
The collected annotations are used only to measure agreement with GPT-based evaluation and are reported only in aggregate form through correlation statistics.

\subsection{Human Alignment Results}
\label{app:human_align_result}

\begin{table}[t]
    \centering
    \small
    \setlength{\tabcolsep}{6pt}
    \begin{tabular}{l c}
        \toprule
        \textbf{QG Dimension} & \textbf{Spearman} \\
        \midrule
        Visual Search Difficulty  & 0.454 \\
        Visual Evidence Coverage  & 0.358 \\
        Visual Context Reasoning  & 0.662 \\
        Visual Spatial Reasoning  & 0.423 \\
        \bottomrule
    \end{tabular}
    \caption{
    Human--GPT evaluation alignment across QG dimensions.
    Spearman correlation is computed between averaged human annotation scores and GPT-based evaluation scores.
    }
    \label{tab:human_gpt_alignment}
\end{table}

Table~\ref{tab:human_gpt_alignment} reports the alignment between averaged human scores and GPT-based evaluation scores. 
Overall, GPT-based evaluation shows positive correlation with human judgments across all four dimensions, suggesting that it captures meaningful trends in human-perceived question quality. 
The strongest alignment is observed for visual context reasoning, with a Spearman correlation of 0.662, indicating that GPT evaluation is relatively consistent with humans when judging whether a question requires higher-level visual interpretation. 
Visual search difficulty and visual spatial reasoning also show moderate positive correlations. 
Visual evidence coverage has the lowest correlation, which is expected because this dimension requires more fine-grained judgment about how broadly a question uses relevant objects, regions, or visual evidence in the image.

These results suggest that our GPT-based evaluator is a reasonable proxy for scalable QG evaluation. 
Although it cannot fully replace human judgment, the positive correlations across all dimensions support its use for comparing model-generated questions at scale.

%% file: Section/Appendix/ExistingVQG_Performance.tex
\section{Comparison with LOVA3}
\label{app:lova3_comparison}

To further contextualize the quality of questions produced by our framework, we compare against LOVA3\citep{zhao2024lova3}, a recent multimodal training paradigm that augments instruction tuning with question asking and VQA-triplet assessment capabilities. 
We use LOVA3-LLaVA-v1.5-7B and compare it with our results based on the Qwen2.5-VL-3B backbone.
LOVA3 relies on existing instruction-tuning data, where supervision is provided by human annotations or stronger external models such as GPT. 
In contrast, our framework constructs self-generated and self-refined supervision from unlabeled images without human annotations or external teacher models.

Since our focus is visual question generation quality, we compare generated questions using the same questioning capability evaluation protocol described in Section~\ref{sec:method_qg_eval}. 
This comparison focuses on perception, reasoning, and diversity dimensions.

\begin{table*}[t]
    \centering
    \small
    \setlength{\tabcolsep}{4.5pt}
    \begin{tabular}{l c c c c c c}
        \toprule
        \textbf{Model} &
        \textbf{Search} &
        \textbf{Coverage} &
        \textbf{Context} &
        \textbf{Spatial} &
        \textbf{Diversity} &
        \textbf{QG Avg.} \\
        \midrule
        LOVA3
        & 0.368 & 0.436 & 0.259 & 0.200 & 0.360
        & 0.325 \\
        Ours, First-round
        & 0.433 & 0.547 & 0.553 & 0.367 & 0.350
        & 0.450 \\
        Ours, Second-round
        & \textbf{0.487} & \textbf{0.607} & \textbf{0.640} & \textbf{0.400} & \textbf{0.370}
        & \textbf{0.501} \\
        \bottomrule
    \end{tabular}
    \caption{
    Comparison with LOVA3 under our visual questioning capability evaluation protocol. 
    We report dimension-level QG scores and the average across all five dimensions.
    }
    \label{tab:appendix_lova3_comparison}
\end{table*}

As shown in Table~\ref{tab:appendix_lova3_comparison}, our first-round model already improves over LOVA3 on visual search difficulty, visual evidence coverage, visual context reasoning, and visual spatial reasoning. 
The largest gains appear on the reasoning dimensions, especially visual context reasoning, where the score increases from 0.259 to 0.553. 
This suggests that our framework produces questions that require more contextual interpretation and spatial reasoning over image evidence.

The second-round model further improves the QG profile and achieves the highest average QG score. 
Compared with LOVA3, the average QG score increases from 0.325 to 0.501. 
These results indicate that our self-evolving framework can generate questions that are more visual-centric and reasoning-demanding under the same evaluation protocol.

%% file: Section/Appendix/Cases.tex
\section{Qualitative Analysis}
\label{sec:exp_qualitative}

\input{Figure_tex/cases}

We provide qualitative examples in Fig~\ref{fig:qualitative_cases} and \ref{fig:qualitative_cases_more} comparing questions generated by the base model $M_0$, the first-round model $M_1$, and the second-round model $M_2$ on the same images.
These examples illustrate how our method gradually improves the model's QG capability.
The base model typically produces simple recognition or attribute-based questions that are valid but require limited visual inspection or reasoning.
After adaptation, the model generates questions that are more grounded in specific visual evidence and involve richer reasoning signals, including cross-region reflection reasoning, relative-size comparison, depth/layout understanding, and fine-grained counting with exclusion.
The second-round model further improves question specificity and relation-centered reasoning, consistent with the quantitative gains observed in visual search difficulty, evidence coverage, visual context reasoning, spatial reasoning, and diversity.

%% file: Figure_tex/cases.tex
\begin{figure*}[t]
  \includegraphics[width=\linewidth]{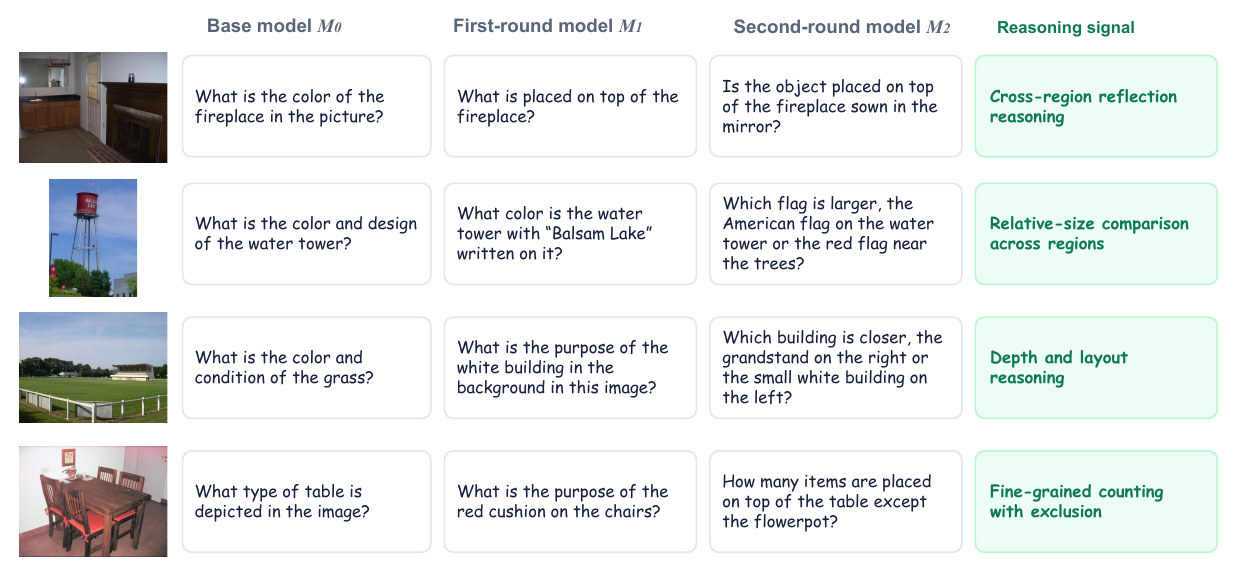}
    \caption{
    \textbf{Qualitative examples of question improvement across generation rounds.}
    We compare questions generated by the base model $M_0$, the first-round model $M_1$, and the second-round model $M_2$ on the same images.
    The base model often asks shallow recognition questions about colors, object types, or scene categories.
    After iterative generation and training, the generated questions become more visually grounded and require stronger reasoning signals, such as cross-region reflection reasoning, relative-size comparison, depth/layout understanding, and fine-grained counting with exclusion.
    }
  \label{fig:qualitative_cases}
\end{figure*}

\begin{figure*}[t]
  \centering
  \includegraphics[width=\linewidth]{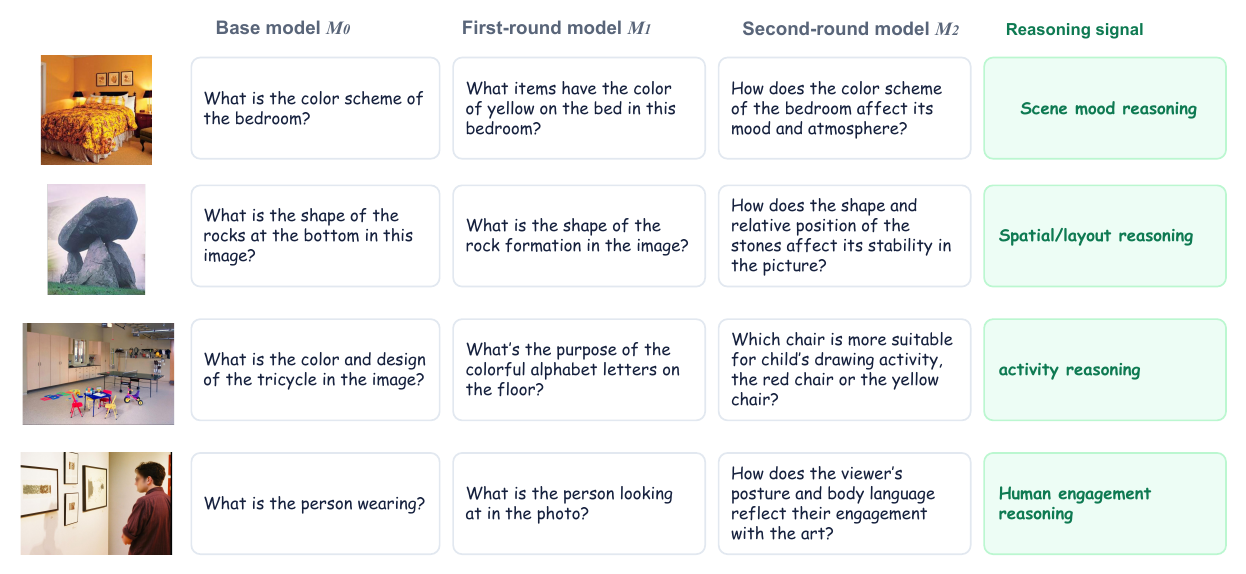}
  \caption{
  \textbf{Qualitative examples of question improvement across generation rounds.}
  We compare questions generated by the base model $M_0$, the first-round model $M_1$, and the second-round model $M_2$ on the same images.
  The base model tends to ask simple recognition questions about color, shape, object type, or visible attributes.
  After iterative generation and training, the questions become more visually grounded and reasoning-oriented, requiring scene mood reasoning, spatial/layout reasoning, activity reasoning, and human engagement reasoning.
  }
  \label{fig:qualitative_cases_more}
\end{figure*}